\documentclass[11pt]{article}

\usepackage{EMNLP2023}
\usepackage{graphicx} 
\usepackage{times}
\usepackage{latexsym}
\usepackage{graphicx}
\usepackage{float}
\usepackage{array}
\usepackage{ulem} 
\usepackage{multirow}
\usepackage{amsmath}
\usepackage{caption}
\usepackage[T1]{fontenc}

\usepackage{tabularx}
\usepackage{microtype}

\usepackage{inconsolata}

\title{SciNLP: A Domain-Specific Benchmark for Full-Text Scientific Entity and Relation Extraction in NLP}

\author{Decheng Duan$^1$\qquad
	Yingyi Zhang$^2$ \qquad
	Jitong Peng$^1$ \qquad
	Chengzhi Zhang$^1$ \footnotemark[2] \\  
	$^1$Nanjing University of Science and Technology, China\\
	$^2$Soochow University, China\\
	\texttt{\{akaddc,pengjit,zhangcz\}@njust.edu.cn},
	\texttt{yyzhang9@suda.edu.cn}
}

\begin{document}
	\pdfoutput=1
	\pagestyle{empty}
	\maketitle
	\renewcommand{\thefootnote}{\fnsymbol{footnote}} 
	\footnotetext[2]{Corresponding authors.}
	\renewcommand{\thefootnote}{\arabic{footnote}}
	\begin{abstract}
		Structured information extraction from scientific literature is crucial for capturing core concepts and emerging trends in specialized fields. While existing datasets aid model development, most focus on specific publication sections due to domain complexity and the high cost of annotating scientific texts. To address this limitation, we introduce SciNLP—a specialized benchmark for full-text entity and relation extraction in the Natural Language Processing (NLP) domain. The dataset comprises 60 manually annotated full-text NLP publications, covering 6,429 entities and 1,649 relations. Compared to existing research, SciNLP is the first dataset providing full-text annotations of entities and their relationships in the NLP domain. To validate the effectiveness of SciNLP, we conducted comparative experiments with similar datasets and evaluated the performance of state-of-the-art supervised models on this dataset. Results reveal varying extraction capabilities of existing models across academic texts of different lengths. Cross-comparisons with existing datasets show that SciNLP achieves significant performance improvements on certain baseline models. Using models trained on SciNLP, we implemented automatic construction of a fine-grained knowledge graph for the NLP domain. Our KG has an average node degree of 3.3 per entity, indicating rich semantic topological information that enhances downstream applications. The dataset is publicly available at: \url{https://github.com/AKADDC/SciNLP}.
	\end{abstract}
	
	\section{Introduction}
	Scientific Named Entity Recognition (SciNER) and Scientific Relation Extraction (SciRE) have emerged as two pivotal technologies for scientific literature mining. SciNER focuses on identifying and categorizing domain-specific scientific entities in publications\citep{9791256,zhang-etal-2021-pdaln,zhong-chen-2021-frustratingly,beltagy-etal-2019-scibert}, while SciRE detects semantic relationships between entities and classifies them into predefined relation types\citep{yan-etal-2021-partition,chen-etal-2020-joint-entity,ji-etal-2020-span,wadden-etal-2019-entity}. These tasks play essential roles in downstream applications such as scientific leaderboard construction, academic Q\&A systems, knowledge graph building, text summarization, and method recommendation\citep{wang-etal-2024-scimon,10.1145/3404835.3462900,10.1016/j.ipm.2023.103574,kedia-etal-2021-beyond-reptile}.
	\begin{figure}
		\centering
		\includegraphics[width=1\linewidth]{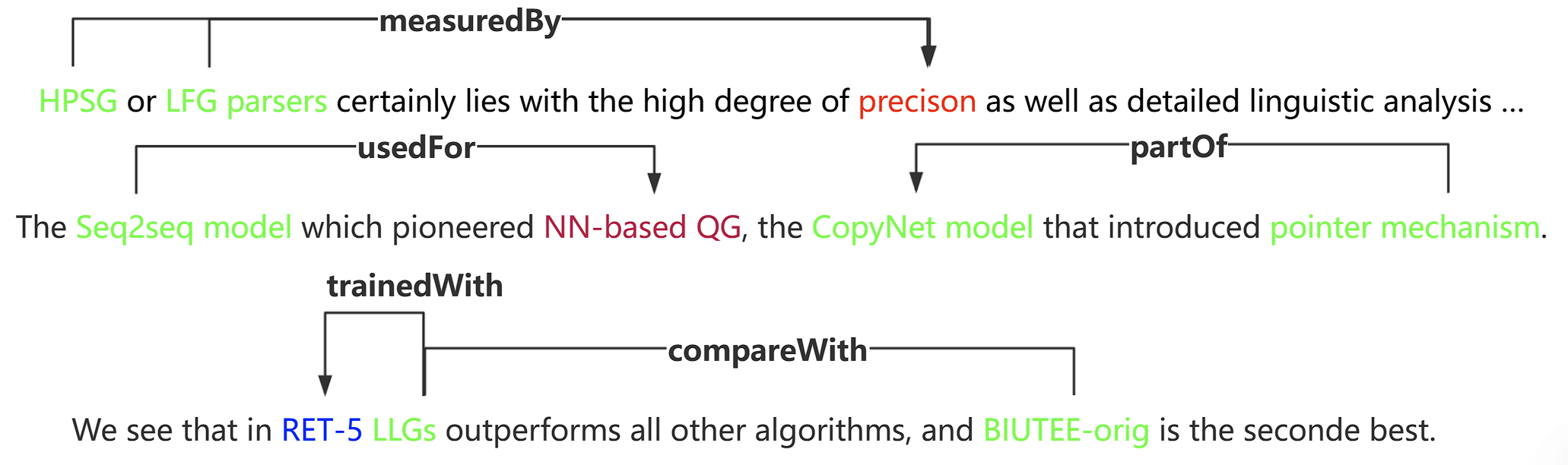}
		\caption{\label{figure1}An annotated example from our dataset, there are four entity types:\textcolor{green}{model},\textcolor{red}{metric},\textcolor{blue}{dataset},\textcolor{purple}{task}, The connecting lines above the sentence represent relationships between entities. In this example, the relationships include measuredBy, usedFor, partOf, trainedWith, and compareWith. Lines with arrows indicate directed relationships, while lines without arrows represent bidirectional relationships.}
	\end{figure}
	Compared to the construction of annotation corpora for traditional Named Entity Recognition (NER) and Relation Extraction (RE) tasks, SciNER and SciRE face significantly greater challenges in corpus development, primarily due to the complexity of scientific texts. First, scientific literature contains a vast array of highly specialized domain-specific terminology, with notable cross-disciplinary variations in terminology systems. Characteristics such as entity overlap and relational complexity require annotators to possess domain expertise. Consequently, high-quality annotated data within a domain typically demands extensive time investment from subject-matter experts. Constructing such corpora for niche disciplines often incurs prohibitively high costs. Second, scientific texts frequently exhibit entity nesting and semantic overlap. Third, scientific relationships demonstrate intricate structural characteristics. They encompass both fundamental method-task associations and complex inter-method relationships . This complexity inherently necessitates heavy reliance on domain knowledge during annotation processes.
	
	We introduce the SciNLP benchmark dataset, the first gold-standard annotated corpus specifically designed for scientific entity and relation extraction tasks in the NLP domain. Figure ~\ref{figure1} illustrates an annotated example from our dataset, demonstrating the predefined complex entity and relation types tailored for the NLP domain. Our dataset distinguishes itself from existing resources through two core differentiating features. 
	
	First, unlike existing scientific corpora that predominantly focus on limited annotation scope to abstracts \citep{10.5555/1289189.1289260,li2016biocreative}, we systematically annotate entities and their relationships across entire scientific papers to ensure contextual completeness during entity extraction—a significantly more challenging task compared to annotating only abstracts or specific sections. This full-text annotation approach enables capturing long-range dependencies and implicit relationships that are crucial for understanding NLP research methodologies. 
	
	Second, while current domain-specific datasets mainly target general AI/machine learning concepts \citep{scidmt,SciERC}, our dataset is specifically designed and annotated for the NLP domain, addressing the critical gap of high-quality annotated resources in NLP, particularly those based on full-text publications. This domain specialization allows precise modeling of NLP-specific research patterns that generic scientific datasets cannot capture. Furthermore, where traditional fine-grained datasets in machine learning or AI typically adopt the simplified TDM framework (Task, Dataset, Method), our dataset introduces more granular entity types (Task, Dataset, Model, and Metric) and richer relation types to capture intricate interactions between method entities. Compared to existing relation schemas that focus on basic comparison or usage relationships, our novel relation taxonomy specifically models the evolutionary dynamics and compositional nature of NLP methods. These relationships, often requiring annotation from main paper sections that previous segment-level annotations neglect, enable deeper semantic characterization of NLP research dynamics, particularly the progression of model architectures and evaluation methodologies over time.
	
	The primary contributions of this work are three folds:
		
		We construct SciNLP, the first manually curated, full-text scientific publication dataset for the NLP domain, comprising 60 ACL conference long papers (2001–2024) with comprehensive semantic annotations. The dataset includes 6,429 fine-grained entities and 1,649 complex relations, addressing the limitations of existing scientific information extraction resources confined to abstracts or narrow subdomains. SciNLP pioneers systematic full-text entity annotation for NLP literature.
		
		A hierarchical ontology framework is developed to reflect NLP’s knowledge structure, defining 4 core entity types (Tasks, Models, Datasets, Metrics) and 11 relation types. This framework captures intricate interactions among NLP’s core entities, enabling precise annotation of the NLP domain’s scientific knowledge.
		
		We validate SciNLP’s effectiveness through experiments on SOTA pipeline and joint extraction models. Models trained on SciNLP demonstrate successful end-to-end joint extraction from unannotated literature, proving the feasibility of constructing fine-grained NLP knowledge graphs.
	\section{Related Work}
	Within the domain of scientific information extraction research, extant scholarly endeavors have established numerous curated datasets for SciNER and SciRE, yet these resources exhibit marked heterogeneity in their annotation schemas and applicability contexts.
	\subsection{Scientific Named Entity Recognition}
	The evolution of scientific literature entity extraction datasets demonstrates a paradigm shift from partial annotation to full-text annotation and from coarse-grained classification to refined taxonomies\citep{li2020survey}. Early initiatives like SEMEVAL-2017\citep{2017-semeval-task-10} concentrated on materials science, annotating three entity types (task/method/material) within selected passages of 500 publications, albeit constrained by limited annotation density and semantic coverage. Recent advancements, exemplified by GSAP-NER\citep{GSAP-NER}, target computer science and machine learning domains, applying 10 fine-grained labels across 100 full-text articles to statistically annotate 54,598 entities, substantially enhancing entity density. The DMDD initiative \citep{DMDD} extends coverage to computer science, NLP, and biomedical fields, combining manual annotation of 450 scientific publications with distant supervision techniques to automatically label an additional 31,219 papers. Furthermore, SciDM\citep{scidmt} establishes three ML-specific entity categories (dataset/method/task), employing a hybrid "precision manual annotation + distant supervision" strategy to annotate 48,149 machine learning full-text articles. This effort yields a corpus of 1.8 million weakly annotated entities, systematically linked to the Paper with Code repository.
	\subsection{Scientific Entity Relation Extraction}
	Furthermore, several datasets concurrently support both entity and relational extraction. Early benchmarks like SEMEVAL-2018\citep{2018-semeval} annotated abstracts in NLP domains, establishing six binary relations including USAGE, though omitting explicit entity type specifications. SCIERC\citep{SciERC} introduced seven scientific relations across 500 AI-focused paper abstracts. Contemporary developments witness SciREX\citep{SciREX} pioneering n-ary relation annotation in ML/NLP domains, albeit without formal n-ary relation definitions, ultimately documenting 360 entities, 169 binary relations, and 5 quaternary relations per paper on average. SciER\citep{scier} represents a paradigm advancement, defining nine relational categories based on SciDMT's framework and annotating 106 full-text AI publications to yield 24,000 entities and 12,000 relations. This progression mirrors the evolutionary trajectory observed in entity extraction datasets, demonstrating parallel developmental patterns between relational and entity annotation initiatives.
	
	Reviewing the extant annotated datasets reveals significant imbalances in disciplinary coverage and annotation depth across current scientific publication resources for SciIE. Cross-domain corpora such as predominantly concentrate on AI and ML domains\citep{SciERC,scidmt,scier}, while specialized disciplinary datasets exhibit marked fragmentation in textual coverage, typically constrained to specific paper sections or isolated sentences\citep{SciREX,2018-semeval}, thereby compromising the integrity of holistic scientific discourse. Addressing these limitations, this study proposes the creation of SciNLP – the inaugural full-text annotation benchmark for NLP research. Anchored in comprehensive semantic units from 60 ACL proceedings (2001-2024), the corpus achieves granular annotation of 6,429 entities and 1,649 relations. Furthermore, it establishes a sophisticated taxonomy of entity and relational categories specifically tailored to the structural and conceptual composition of NLP scholarly articles.
	
	\section{SciNLP Corpus}
	This section we delineate the methodological framework underlying the SciNLP corpus construction, including Data Curation and Preprocessing Pipeline in ~\ref{3.1}, Annotation Protocol and Quality Assurance Mechanisms in ~\ref{3.2}, and Comparative Analysis with Benchmark Corpora in ~\ref{3.3}.
	\subsection{Data Collection and Pre-processing}
	\label{3.1}
	Our data collection began by integrating the ACL OCL Corpus¹\footnotetext[1]{https://github.com/shauryr/ACL-anthology-corpus}, which aggregates full-text papers and metadata from the ACL Anthology platform²\footnotetext[2]{https://aclanthology.org/} up to September 2022, totaling 73,285 papers. To address the corpus’s update lag, we additionally acquired 9,387 newly published papers (October 2022 to December 2024) from the ACL Anthology, extracting their PDF content and aligning metadata using the document parsing tool $GROBID^3$\footnotetext[3]{https://github.com/kermitt2/grobid}, resulting in a total collection of 82,672 papers.Given the academic authority and technical cutting-edge nature of ACL Long Papers (Association for Computational Linguistics Long Papers)—which best represent NLP’s technological evolution—we applied stratified sampling principles to randomly select 60 papers from the full-text corpus of ACL Long Papers (2001–2024) as our annotated sample set (see Appendix ~\ref{appendix} for the specific sampling strategy).
	\begin{table}
		\centering
		\small
		\begin{tabular}{lccc} 
			\hline
			\textbf{~}  \textbf{Items} & \textbf{SemEval18} & \textbf{TDMSci} & \textbf{SciNLP}     \\ 
			\hline
			{Annotation range} & Abstract           & Sentence        & Full text  \\
			Entity types                                   & -                  & 3               & 4          \\
			Relationship~types                             & 6                  & -               & 11         \\
			Entities                                       & -                  & 2937            & 6409       \\
			Relationship                                   & 1595               & -               & 1648       \\
			Documents                                      & 500                & -               & 60         \\
			\hline
		\end{tabular}
		\caption{\label{table1}Comparison of SciNLP and 2 datasets in NLP domain, the "-" indicates that there is no corresponding data in the dataset.}
	\end{table}
	\subsection{Comparison with Previous Datasets}
	\label{3.2}
	\textbf{Annotation Scheme}: We constructs an information extraction annotation framework tailored to the structured characteristics of academic literature in the NLP domain. Specifically, we define four representative entity types for NLP publications—Task, Model, Dataset, and Metric—by synthesizing domain-specific requirements and prior knowledge entity classifications for NLP\citep{qasemizadeh-schumann-2016-acl}. These entity types are prevalent in core sections of NLP papers, such as methodology and experimental design. Annotation guidelines explicitly instruct annotators to label only factual entities with concrete meanings, ensuring consistency and relevance.You can see Appendix ~\ref{appendix} for the specific annotation scheme.
	
	\begin{table*}
		\centering
		\fontsize{10}{10}\selectfont
		\renewcommand{\arraystretch}{1.5}
		\begin{tabular}{>{\hspace{0pt}}m{0.137\linewidth}>{\hspace{0pt}}m{0.363\linewidth}>{\hspace{0pt}}m{0.381\linewidth}}
			\hline
			\multicolumn{1}{c}{\textbf{Relation type}}& \multicolumn{1}{c}{\textbf{Examples}} &\multicolumn{1}{c}{\textbf{Example sentences}}                                     \\ 
			\hline
			measuredBy      & \textcolor{orange}{HPSG}\(\to\)\textcolor{red}{precision}     &\textcolor{orange}{HPSG} or LFG parsers certainly lies with the high degree of \textcolor{red}{precision} as well as detailed linguistic analysis …  \\

			evaluatedBy & \textcolor{purple}{SQG}\(\to\)\textcolor{blue}{CoQA} & Prior works performed \textcolor{purple}{SQG} on \textcolor{blue}{CoQA}\\
			
			usedFor   & \textcolor{orange}{Seq2Seq architecture}\(\to\)\textcolor{purple}{NN-based QG} & Du pioneered \textcolor{purple}{NN-based QG} by adopting the \textcolor{orange}{Seq2Seq architecture}… \\
			
			enhancedBy   & \textcolor{orange}{Seq2Seq modek}\(\to\)\textcolor{orange}{variational inference}    & Enhancing the \textcolor{orange}{Seq2Seq modek} into complicated strutures using \textcolor{orange}{variational inference}…\\
			
			evaluatedOn   & \textcolor{purple}{QG}\(\to\)\textcolor{blue}{SQuAD} ; \textcolor{purple}{QG}\(\to\)\textcolor{blue}{NewsQA}    & \textcolor{purple}{QG} is ofen performed on existing QA datasets , e.g. , \textcolor{blue}{SQuAD} , \textcolor{blue}{NewsQA} , etc.\\
			
			partOf   & \textcolor{orange}{self-attention}\(\to\)\textcolor{orange}{multi-head attention}    & In the \textcolor{orange}{multi-head attention} layer , there are not only lh vanilla \textcolor{orange}{self-attention} …\\
			
			compareWith   & \textcolor{orange}{CorefNet}---\textcolor{orange}{ReDR}    & Besides , \textcolor{orange}{CorefNet} gets better performance among all baselines , especially \textcolor{orange}{ReDR}\\
			
			subtaskOf   & \textcolor{purple}{SQG}\(\to\)\textcolor{purple}{conversation generation}    & Second , since prior works regarded \textcolor{purple}{SQG} as a \textcolor{purple}{conversation generation}, …\\    
			
			similarWith   & \textcolor{orange}{Boltzmann-machine}---\textcolor{orange}{BM}    & We propose to adopt the Boltzmann-machine (BM) distribution as a variational…\\   
			
			trainedWith   & \textcolor{orange}{BERT}\(\to\)\textcolor{blue}{Persona-Chat} ; \textcolor{orange}{GPT-2}\(\to\)\textcolor{blue}{Persona-Chat}    & Group 1 applies direct fine-tuning of \textcolor{orange}{BERT} or \textcolor{orange}{GPT-2} on \textcolor{blue}{Persona-Chat}…\\ 
			
			subclassOf   & \textcolor{orange}{RPART}\(\to\)\textcolor{orange}{CART}    & \textcolor{orange}{RPART} , which is a \textcolor{orange}{CART} reimplementation for the S-Plus and R statistical computing enviroments…\\ 
			\hline
		\end{tabular}
		\caption{\label{table2}Examples of the relations types, the four entity types: \textcolor{orange}{Model}, \textcolor{red}{Metric}, \textcolor{blue}{Dataset}, \textcolor{purple}{Task}.}
		
	\end{table*}
	For relation annotation, building on prior work in ML, AI, and NLP domains\citep{SciERC,scier} and aligning with the structural conventions of NLP literature, we define 11 relation types to represent complex interactions among the four entity categories: MeasuredBy, UsedFor, TrainedWith, EnhancedBy, SubclassOf, SubtaskOf, PartOf, EvaluatedBy, EvaluatedOn, SimilarWith, and CompareWith. Excluding the symmetric relations SimilarWith and CompareWith, all other relation types are directional. Table ~\ref{table2} provides some annotation examples of these relations. Compared to existing studies, our framework offers finer-grained distinctions, particularly for interactions between Model entities, where we employ three asymmetric relations—SubclassOf, EnhancedBy, and PartOf—to precisely characterize hierarchical, enhancement, and compositional dependencies. Detailed specifications of entity-relation labels are provided in Appendix ~\ref{appendix}.
	
	\textbf{Annotation Strategy}:
	We employs $Doccano^4$\footnotetext[4]{https://github.com/doccano/doccano} as the annotation tool, with a two-stage annotation process. First, the 60 papers are evenly divided into five chronological groups (12 papers per group). Two NLP experts conducted a pilot annotation on one group: they drafted initial entity annotation guidelines, independently annotated the papers, then resolved discrepancies through discussions to refine the guidelines. These optimized guidelines were used to train five NLP graduate students for subsequent annotations. The remaining four groups were assigned to the students, with each paper annotated by at least two annotators. Inter-annotator agreement analysis revealed Cohen’s kappa scores (Andres et al.) of 0.90, 0.91, 0.83, and 0.86 for entity annotation across the four groups, and 0.75, 0.81, 0.71, and 0.64 for relation annotation, demonstrating reliable annotation consistency.
	\subsection{Data Analysis}
	\label{3.3}
	Upon completing the annotation process, our dataset comprises 6,429 fine-grained entities and 1,649 relations, with an average of 106 entities and 27 relations per document. As shown in Table~\ref{table1}, the scale of our dataset substantially exceeds prior NLP-specific datasets designed for entity and relation extraction. The annotated data was then randomly shuffled and split into training, validation, and testing sets in an 17:2:1 ratio to support subsequent experimental analyses.\footnotemark[5]\footnotetext[5]{When constructing the data subsets, we partitioned the dataset into training, validation, and test sets at an 17:2:1 ratio based on the quantities of entities and relations, thereby ensuring consistency in the distribution of data characteristics across all subsets.}
	\section{Document-Level Entity and Relation Extraction}
	This section primarily elaborates on our definition of the full-text-based Entity and Relation Extraction (ERE) task (Section ~\ref{4.1}) and details the evaluation metrics adopted for assessment (Section ~\ref{4.2}).
	\subsection{Task Definition}
	\label{4.1}
	The end-to-end Entity and Relation Extraction (ERE) task aims to automatically extract structured triplets T = {($e_{i}$, $r_{k}$, $e_{j}$)} from scientific literature D, where each triplet represents two scientific entities ($e_{i}$, $e_{j}$) and their semantic relation $r_{k}$. Current research typically decouples this task into two core subtasks: SciNER (Scientific Named Entity Recognition) and SciRE (Scientific Relation Extraction). Existing methods primarily follow two technical paradigms: (1) Pipeline approaches sequentially execute SciNER and SciRE—first identifying entity boundaries/types, then classifying relation types between entities; (2) Joint learning frameworks synchronously optimize both subtasks through parameter sharing or unified annotation strategies, effectively mitigating error propagation issues inherent in traditional pipeline methods.
	
	As our dataset is annotated at the full-text level, the basic data unit is document-based. For modeling, we input a scientific document D, which comprises multiple sentences P = {$p_{1}$, $p_{2}$, ..., $p_{n}$}, each composed of word sequences S = {$w_{1}$, $w_{2}$, ..., $w_{m}$}. Based on this structure, we formally define the SciNER and SciRE tasks as follows:
	
	\textbf{SciNER}: For each predefined entity type E, the SciNER model identifies all possible entities in every sentence of an input document. It detects the span $Span_{i}${$w_{l}$, $w_{r}$} and predicts the entity type $c_{i} \in E$ for each entity, where $w_{l}$ denotes the leftmost index and $w_{r}$ the rightmost index of the entity span within the word sequence W = {$w_{1}$, $w_{2}$, ..., $w_{n}$}.
	
	\textbf{SciRE}: Given a predefined set of relation types R, the SciRE model predicts whether a relation $r_{i} \in R$ exists between every pair of entities ($e_{i}$, $e_{j}$) within the same sentence P, where entities are drawn from the set of entity mentions Entity.
	\subsection{Evaluation Settings and Results}
	\label{4.2}
	To systematically evaluate the performance differences between joint Entity and Relation Extraction (ERE) and pipeline approaches, this study establishes granular evaluation frameworks for each subtask:
	
	\textbf{Named Entity Recognition (NER) Evaluation:}
	Adopts strict span matching criteria, requiring models to correctly identify both entity boundaries and semantic types.
	
	\textbf{End-to-End Relation Extraction (End-to-End RE) Evaluation}:For the ERE extraction results, we add two additional indicators Rel and Rel+ to evaluate the accuracy.
	
	\textbf{Boundary-Aware Metric (Rel)}: Following prior work\citep{PURE,pl-marker,hgere}, a relation is considered correct only if:
	(a) Subject entity boundaries are correctly identified;
	(b) Object entity boundaries are correctly identified;
	(c) Relation type is accurately predicted.
	
	\textbf{Strict Metric (Rel+)}: Extends Rel by additionally requiring correct prediction of subject/object entity types.
	
	\textbf{Pure Relation Classification (RE) Evaluation}:Evaluates models’ ability to discern semantic relationships between entities under gold-standard entity annotations. Specifically, for any entity pair:
	If a predefined relation exists, the model must predict its type correctly;
	If no relation exists, the model must correctly identify the absence of a relation.
	
	\begin{table*}
		\centering
		\small
		\begin{tabular}{lcccccccc} 
			\hline
			\multirow{2}{*}{\textbf{Methods}} & \multicolumn{4}{c}{\textbf{Sentence-level Test}} & \multicolumn{4}{c}{\textbf{Doc-level Test}}   \\ 
			\cline{2-9}
			& \textbf{NER}   & \textbf{Rel}   & \textbf{Rel+}  & \textbf{RE}        & \textbf{NER}   & \textbf{Rel}   & \textbf{Rel+}  & \textbf{RE}     \\ 
			\hline
			PURE\citep{PURE}                         & 59.42          & 27.53          & 26.96          & 28.35              & 62.86       & 29.93        & 29.46         & {29.17}    \\
			PL-Marker\citep{pl-marker}                        & 60.84          & {29.06}          & {28.41}          & \textbf{34.87}     & 63.72          & {32.09}          & {30.73}          & \textbf{36.87}  \\
			HGERE\citep{hgere}                           & \textbf{78.25} & \textbf{45.56} & \textbf{44.18} & -                  & \textbf{79.53} & \textbf{49.28} & \textbf{47.64} & -               \\
			\hline
		\end{tabular}
		\caption{\label{table3}Comparison F1 scores between full text and sentence on test set.  “Rel” and “Rel+” denote the results of end-to-end relation extraction under boundaries evaluation and strict evaluation, respectively. “RE” indicates performing relation extraction with given gold standard entities, applicable only to pipeline extraction methods. "-" indicates that this model is not capable of handling the task. }
	\end{table*}

	\begin{table*}
		\centering
		\small
		\begin{tabular}{lcccc|cccc|cccc} 
			\hline
			\multirow{2}{*}{\textbf{Methods}} & \multicolumn{4}{c}{\textbf{SciNLP}} & \multicolumn{4}{c}{\textbf{SCIERC}} & \multicolumn{4}{c}{\textbf{SciER}}  \\ 
			\cline{2-13}
			& \textbf{NER} & \textbf{Rel} & \textbf{Rel+} & \textbf{RE}  & \textbf{NER} & \textbf{Rel} & \textbf{Rel+} & \textbf{RE}  & \textbf{NER} & \textbf{Rel} & \textbf{Rel+} & \textbf{RE}  \\ 
			\hline
			PURE                                 & 62.86       & 29.93        & 29.46         & 29.17        & 66.68        & 48.65        & 35.68         & 37.54        & 81.60        & 53.27        & 52.67         & 73.99        \\
			PL-Marker                                    & 63.72          & {32.09}          & {30.73}          & 36.87          & 69.94        & 53.25        & 41.62         & 43.05        & 83.31        & 60.06        & 59.24         & 77.11        \\
			HGERE                                  & 79.53        & 49.28        & 47.64         & -            & 74.91        & 55.72        & 43.6          & -            & 86.85        & 62.32        & 61.10         & -            \\
			\hline
		\end{tabular}
		\caption{\label{table4}F1 scores of different baseline models on the SciNLP, SciERC, and SciER datasets. The baseline results for SciERC and SciER are cited from \citep{hgere}.}
	\end{table*}

	\section{Experiments}
	In this section, we utilize SciNLP to train SOTA supervised methods for SciNER and SciRE tasks, and compare them with existing datasets to validate the efficacy of our dataset. Section ~\ref{5.1} introduces the supervised Baselines we have used. Section ~\ref{5.2} presents the results of our dataset across these methods, followed by detailed discussions and analyses.
	\subsection{Supervised Baselines}
	\label{5.1}
	To validate the effectiveness of our dataset, we selected state-of-the-art (SOTA) methods for SciNER and SciRE tasks, including the following three approaches:
	
	\textbf{PURE}: \citep{PURE} This model employs a dual-encoder architecture to decouple entity recognition and relation extraction tasks. The entity model performs span-based type prediction, while the relation model enhances classification through dynamic entity pair inputs and the insertion of entity-type markers. It introduces crosssentence context expansion by extending input windows, leveraging pre-trained models’ ability to capture long-range dependencies for improved global semantic reasoning.
	
	\textbf{PL-Marker}: \citep{pl-marker} Proposing a floating marker mechanism, this approach dynamically binds markers to text spans during the entity recognition phase through shared positional encoding and boundary-aware attention alignment, with neighborhood clustering further optimizing boundary learning. In the relation extraction phase, it restructures spans and markers via subject-centric reorganization, jointly encoding subject-object semantics to overcome limitations of independent span-pair modeling. This unified framework enables seamless integration of entity and relation representations.
	
	\textbf{HGERE}: \citep{hgere} Replacing strict NER modules with a lightweight pruning module, this method generates high-recall candidate entities, transferring classification tasks to a joint module to mitigate error propagation. It constructs hypergraph structures (comprising entity nodes, relation nodes, and multi-dimensional hyperedges) to aggregate high-order semantic information through hierarchical message passing. This design significantly outperforms traditional GNNs limited to pairwise interactions, particularly in capturing complex scientific logic across long documents.
	
	All methods were implemented using the SciBERT-scivocab-uncased pre-trained model as the text encoder. For fair comparision, experiments adopted identical random seed sets to ensure robustness, along with consistent fine-tuning settings: learning rate = 2e-5, weight decay = 0.01, batch size = 16, and context length = 100.
	
	
	
	\subsection{Result Analysis}
	\label{5.2}
	For task evaluation, we assessed the performance of end-to-end models on each subtask and conducted ablation experiments to verify whether full-text annotation improves extraction performance compared to segment-specific approaches. Specifically, we processed SciNLP into two variants:
	\textbf{Full-Text Annotation}: Retains the complete document content, providing models with rich contextual information.
	\textbf{Segment-Restricted Annotation}: Extracts only sentences containing entities and relations, forming a sentence-level dataset with the context window length set to 0 to isolate local semantic dependencies.
	As demonstrated in Table 3, models trained on the full-text variant achieved superior performance in capturing cross-paragraph scientific logic and fine-grained relationships, validating the necessity of full-text annotation for robust scientific information extraction.
	
	Systematic analysis of the ablation study results in Table ~\ref{table3} reveals the following key findings: First, in document-level context modeling, all models exhibit significantly superior performance on the full-text annotated dataset compared to sentence-level settings, with SciRE demonstrating the most notable improvement. For instance, the PL-Marker model achieves a 2.32 percentage point gain on the Rel+ task under document-level training versus sentence-level training, while RE metrics also show enhancements, underscoring the critical role of cross-sentence contextual information in supporting complex relation reasoning. Notably, during relation-enhanced tasks (Rel+), document-level training ensures stable performance gains across models, with HGERE exhibiting the largest improvement due to the advanced capability of its dynamic graph attention mechanism in integrating global semantic information. These experimental results empirically validate that optimizing scientific information extraction models relies not only on algorithmic innovation but also on synergistic evolution with contextually complete annotation frameworks, where data richness and architectural design mutually reinforce performance.
	
	Document-level context consistently improves all three architectures, with particularly stable gains on relation extraction. Meanwhile, model comparisons reveal that HGERE obtains an unusually large advantage over pipeline baselines on SciNLP, indicating that SciNLP provides a challenging and discriminative benchmark for joint entity–relation modeling. Table ~\ref{table4} presents a comparative evaluation of different baseline models on three scientific information extraction datasets: SciNLP, SCIERC, and SciER, revealing notable interaction effects between dataset characteristics and model architectures.
	
	First, for the named entity recognition (NER) task, all three datasets exhibit a consistent performance improvement as model complexity increases. Meanwhile, the overall NER performance on SciNLP remains lower than that on SciER but is comparable to SCIERC, suggesting that entity recognition performance is still largely influenced by annotation scale and domain consistency.
	
	Second, more pronounced performance disparities emerge in relation extraction–related tasks. Overall, SciER consistently achieves superior performance on both the Rel and Rel+ tasks compared to SciNLP and SCIERC, which can be attributed to its higher relation annotation density and more mature dataset construction paradigm. Notably, on the SciNLP dataset, HGERE consistently achieves the best NER, Rel, and Rel+ performance across all three datasets. Notably, its margin over PL-Marker is substantially larger on SciNLP than on SciERC or SciER, suggesting that SciNLP may provide a more discriminative setting for evaluating joint higher-order entity–relation modeling.
	
	In summary, the results indicate that SciNLP remains challenging in terms of absolute relation extraction performance, while also producing substantially larger performance differences between pipeline and joint higher-order models. This suggests that SciNLP may serve as a discriminative benchmark for evaluating architectural advances in joint scientific entity–relation extraction.
	\begin{figure*}
		\small
		\centering
		\includegraphics[width=1\linewidth]{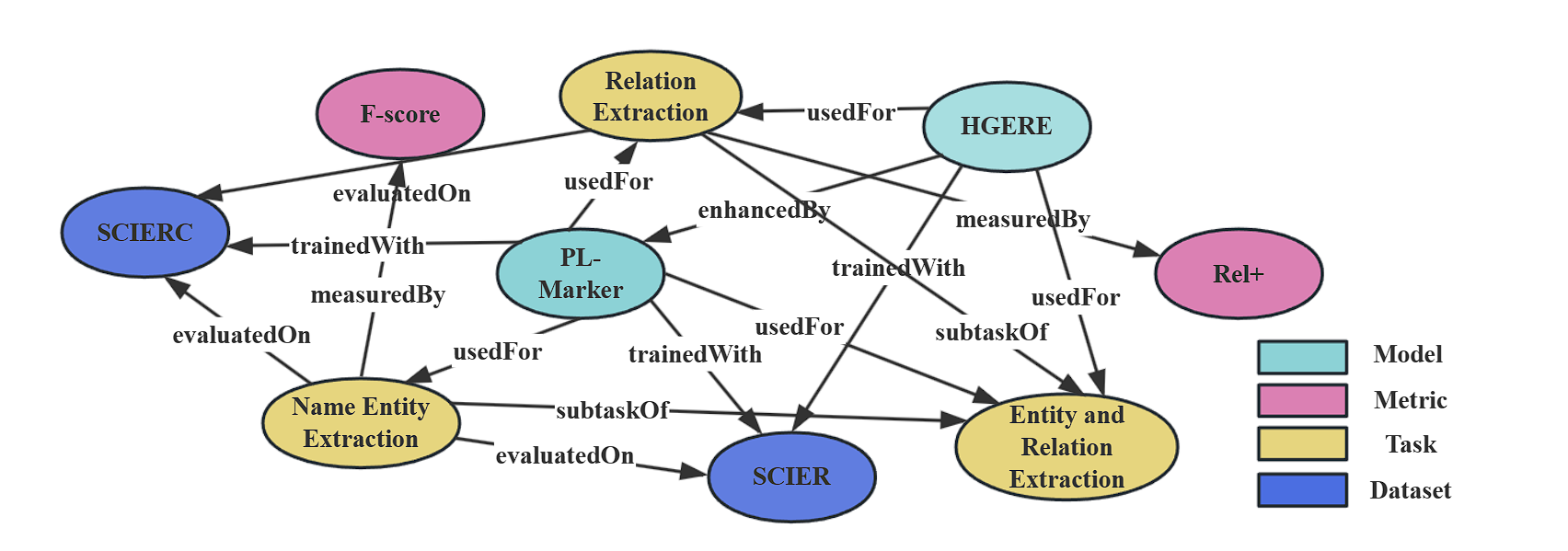}
		\caption{\label{figure2}A subset of the NLP fine-grained KG}
	\end{figure*}

	\section{Building Fine-grained KG in NLP Domain}
	
	We construct a scientific knowledge graph from the entire ACL corpus. First, we built a comprehensive corpus containing full texts and metadata of all ACL conference papers (2001–2024) from the ACL Anthology repository, totaling 82,672 papers. Leveraging the HGERE and PL-Marker models, we performed end-to-end triplet extraction across all full-text papers to build a fine-grained NLP knowledge graph, where nodes represent extracted scientific entities and edges denote semantic relationships between entities. This framework enables systematic encoding of domain-specific knowledge structures and cross-document scientific logic.
	
	The construction of our knowledge graph involves two key steps:
	First, the model extracts all triplet data T from each paper, preserving every triplet in the form ($e_{i}$, $r_{k}$, $e_{j}$), where $e_{i}$ and $e_{j}$ are related entities, and $r_{k}$ denotes their semantic relationship. Second, we perform entity normalization to ensure uniqueness within the knowledge graph. This process begins by creating an abbreviation-full form mapping table between entities using the predefined similar relationship. Subsequently, we unify entities using the entity normalization method proposed by\citep{10.1016/j.ipm.2023.103574}, which calculates similarity scores based on lexical features and applies clustering algorithms to standardize entity representations. This dual-step approach guarantees both the richness and semantic consistency of the knowledge graph. In the end, we constructed a fine-grained NLP knowledge graph comprising 205k nodes and 693k edges, with an average node degree of 3.3. Figure ~\ref{figure2} illustrates a subset of the constructed graph, demonstrating its ability to effectively preserve domain-specific knowledge in NLP. This topology reflects the intricate semantic networks inherent to computational linguistics research.
	
	Table ~\ref{table6} presents the statistics of nodes and relations in the KG we constructed. It can be observed that in the distribution of entity type quantities, entities of the "model" type account for a relatively high proportion. This indicates that NLP domain papers demonstrate a significant model-centric research paradigm, reflecting the core characteristics of rapid technological iteration and model diversification in this field. Additionally, the predominant occurrence of "usedFor" and "measuredBy" relationships highlights the characteristic of method and model-driven technological development in the NLP domain.
	
	\begin{table}[ht]
		\centering
		\small
		\begin{tabular}{>{\hspace{0pt}}m{0.15\linewidth}>{\hspace{0pt}}m{0.15\linewidth}>{\hspace{0pt}}m{0.01\linewidth}>{\hspace{0pt}}m{0.25\linewidth}>{\hspace{0pt}}m{0.15\linewidth}} 
			\hline
			\multicolumn{2}{c}{{\textbf{Nodes}}} &  & \multicolumn{2}{c}{\textbf{Relations}}  \\ 
			\cline{1-2}\cline{4-5}
			\textbf{Type}  & \textbf{Count}                                    &  & \textbf{Type}  & \textbf{Count}                                         \\ 
			\hline
			Task           & 33,684                                            &  & measuredBy     & 117,050                                                \\
			Dataset        & 31,587                                            &  & evaluatedBy    & 24,325                                                 \\
			Metric         & 30,751                                            &  & usedFor        & 171,930                                                \\
			Model          & 109,636                                           &  & enhancedBy     & 32,329                                                 \\ 
			&                                                   &  & evaluatedOn    & 37,742                                                 \\ 
			&                                                   &  & partOf         & 44,462                                                 \\ 
			&                                                   &  & compareWith    & 77,039                                                 \\ 
			&                                                   &  & subtaskOf      & 40,454                                                 \\ 
			&                                                   &  & similarWith    & 37,951                                                 \\ 
			&                                                   &  & trainedWith    & 91,147                                                 \\ 
			&                                                   &  & subclassOf     & 18,822                                                 \\ 
			\hline
			\textbf{Total} & 205,658                                           &  & \textbf{Total} & 693,251                                                \\
			\hline
		\end{tabular}
		\caption{Nodes and relations statistics}
		\label{table6}
	\end{table}
	
	\section{Conclusion}
	We present SciNLP, a dataset specifically designed for fine-grained entity and relation extraction in the Natural Language Processing (NLP) domain. The dataset defines four entity types (Task, Model, Dataset, Metric) and 11 relation types (MeasuredBy, UsedFor, TrainedWith, EnhancedBy, SubclassOf, SubtaskOf, PartOf, EvaluatedBy, EvaluatedOn, SimilarWith, CompareWith), overcoming the limitations of coarse-grained relation definitions in existing resources. Constructed from 60 ACL conference papers spanning 2001–2024, SciNLP contains 6,429 entities and 1,649 relations. Unlike cross-disciplinary or abstract-level datasets, SciNLP preserves full-text scientific context through document-level annotation, providing critical support for models to capture cross-paragraph semantic dependencies.
	Experimental results demonstrate that models trained on SciNLP significantly outperform sentence-level settings in document-context modeling. Leveraging SciNLP, we automatically extracted knowledge from 82,672 NLP publications, constructing a fine-grained knowledge graph with 205k nodes and 693k edges, where entities exhibit an average node degree of 3.3. SciNLP establishes a high-confidence benchmark for NLP knowledge mining, with its full-text annotation paradigm and fine-grained ontology design enabling downstream applications such as domain-specific QA and technological trend analysis. 
	
	\section*{Limitations}
	While SciNLP holds pioneering value as the first full-text annotated dataset for the NLP domain, it exhibits several limitations: First, in terms of dataset scale, SciNLP’s entity and relation coverage is smaller compared to cross-domain resources like SciER. Second, it does not address the widespread phenomenon of nested entities in scientific texts. Third, structural parsing errors in some full-text papers (via Grobid) introduced annotation inconsistencies. Finally, the constructed knowledge graph lacks integration with SOTA KG construction methods, resulting in suboptimal entity disambiguation and alignment rates. These gaps highlight directions for future improvements, such as incorporating nested entity modeling and robust cross-document linking frameworks.
	
	\section*{Ethical Statement}
	Our \textbf{SciNLP} dataset is constructed based on resources from The ACL OCL Corpus\footnotemark[6]\footnotetext[6]{https://github.com/shauryr/ACL-anthology-corpus}, a large corpus which provides full-text and metadata to the ACL anthology collection, and it is released under the CC BY-NC 4.0\footnotemark[7]\footnotetext[7]{https://creativecommons.org/licenses/by-nc/4.0/deed.en}. All the other data are public from scientific documents. We release dataset for scientific information extraction tasks. There are no risks in our work. 
	
	In the process of the dataset construction, all human annotators are from our research group IR\&TM, and all members participate voluntarily. The SciNLP dataset includes metadata extracted from the papers. No sensitive personal data (e.g., contact details or affiliations) is included. All metadata was collected in compliance with the terms of their sources and is used strictly for noncommercial academic research. We are committed to handling the data responsibly and ethically and will release our dataset under the same non-commercial license to ensure transparency and responsible data usage.
	
	\section*{Acknowledgements}
	This work is supported by the National Natural Science Foundation of China (No.72074113). We gratefully acknowledge all the members of the IR\&TM team who participated in the annotation of data sets. We thank Yuxin Xie, Heng Zhang for providing the basic data sets and the related codes.

	\bibliography{anthology,custom}
	\bibliographystyle{acl_natbib}
	\clearpage
	
	\newpage
	\appendix
	
	\section{Annotation Guideline}
	\label{appendix}
	This section outlines the annotation process for identifying entities and relationships in NLP literature. The workflow includes the use of annotation tools, the design of the annotation process, the definition of the entity types and relation types, and concrete examples of annotated content, which will be detailed sequentially below.
	
	\subsection{Annotation Tool}
\begin{figure}
		\centering
		\includegraphics[width=1\linewidth]{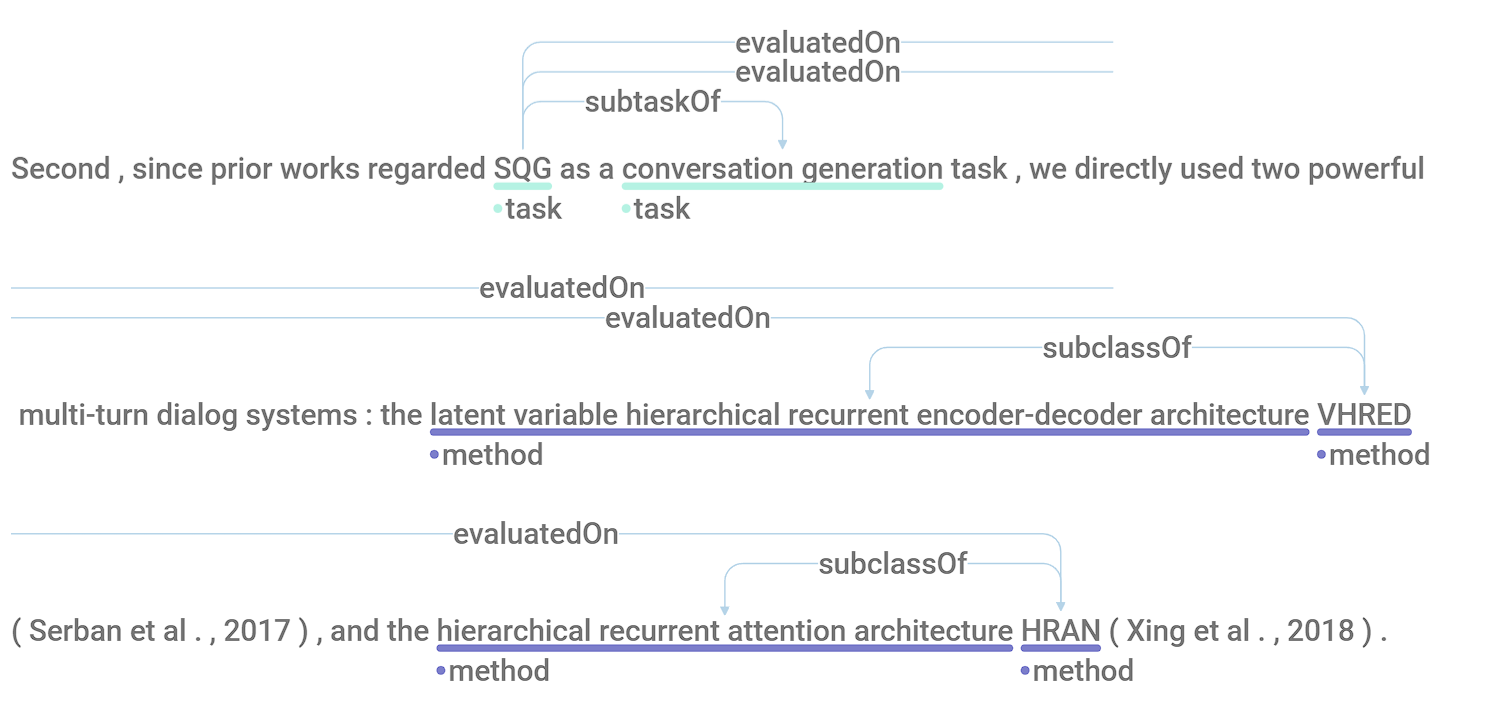}
		\caption{\label{figure3}Doccano Annotation interface}
	\end{figure}

	This data annotation task focuses on four types of entities in full-text NLP academic papers: algorithm entities, dataset entities, metric entities, and task entities. The original full-text corpus is sourced from the ACL Anthology Reference Corpus, where papers have been converted from XML/PDF formats to TXT format. For this study, we randomly selected 60 papers to build the training set for the automatic method of entity extraction. The manual annotation of method/task entities in these 60 papers was conducted using the online annotation platform Doccano.Figure ~\ref{figure3} shows our annotation interface.

	\subsection{Entity Annotation}
	\textbf{A.2.1 Entity Types}
	
	We annotate the following four entity types, Table ~\ref{table7} shows examples of correct entity annotations:
	\begin{itemize}
		\item \textbf{Task}: A task constitutes an objective requiring operational resolution through systematic processes.(e.g., Text Generation, Text Classification, Sentiment Analysis, Named Entity Recognition, and Automatic Keywords Extraction).
		\item \textbf{Model}: Model entities refer to specific algorithms or model architectures explicitly mentioned in scientific literature or technical documentation, designed to solve computational tasks.(e.g., GPT, LDA, SVM, LSTM, Skip-gram, Adam, BERT).
		\item \textbf{Dataset}: Dataset entities represent structured data resources explicitly used or referenced in scientific research for training, evaluation, or analysis. (e.g., Brown Corpus, Penn Treebank, WordNet).
		\item \textbf{Metric}: Metric entities denote quantitative measures used to evaluate the performance of models, algorithms, or experimental results in scientific research. (e.g., Accuracy, Precision, Recall, F1-score, BLEU, ROUGE, Kappa).
	\end{itemize}
	
	\textbf{A.2.2 Annotation Note}
	\begin{itemize}
		\item Determiners (e.g., "a," "the," "some," "our") placed before nouns for grammatical purposes should not be included in the entity span. For example, in the phrase "the FB15K dataset," only "FB15K dataset" should be annotated as the resource entity. However, if it is a comparison between models, it is necessary to judge whether to retain the modifiers before and after the entity according to the context.
		\item Annotators must mark the smallest span necessary to represent the core meaning of an algorithm/tool/resource/metric entity. Non-essential adjectival modifiers (e.g., adjectives) that are not integral to the entity’s definition should be excluded. For instance, in "novel BERT-based model," only "BERT-based model" would be annotated (omitting "novel" if it is not part of the formal name).
		\item Do not annotate anonymous entities that include anaphoric expressions (e.g., "this algorithm," "this metric," or "the dataset"). Only annotate "factual, content-bearing" entities. Algorithm/tool/resource/metric entities typically have specific names and maintain consistent meanings across different papers. These entities are usually single terms, phrases, or abbreviations. However, even if an entity is "factual and content-bearing," it should not be annotated if its meaning is confined to a single paper’s specific research topic or problem (e.g., an ad-hoc method named only within that paper’s context).
		\item If a method/algorithm entity ends with terms such as "algorithm," "model," "framework," "strategy," "dataset," "corpus," or "toolkit," annotate the entire span, including the suffix. For example, for the algorithm "Support Vector Machine (SVM)," if it appears as "SVM algorithm" in a paper, annotate the full term "SVM algorithm." If it appears as "SVM" alone without a suffix indicator, annotate only "SVM." This ensures consistency in capturing formal designations while adhering to the minimal span principle when no clarifying suffix is present.
		\item Generic terms for method categories (e.g., "classification algorithm," "semi-supervised method," "machine learning," or "deep learning") should not be annotated. Only annotate specific, named instances with well-defined meanings, such as concrete algorithm entities (e.g., SVM, KNN, Decision Trees). This ensures that annotations focus on identifiable, context-independent entities rather than broad conceptual categories.
	\end{itemize}
	
	\begin{table*}
		\centering
		\renewcommand{\arraystretch}{1.5}
		\begin{tabular}{>{\hspace{0pt}}m{0.1\linewidth}>{\hspace{0pt}}m{0.5\linewidth}>{\hspace{0pt}}m{0.1\linewidth}>{\hspace{0pt}}m{0.2\linewidth}} 
			\hline
			\textbf{Examlpe} & \textbf{Sentence}                                                                                                                                                                                                        & \textbf{Entity type} & \textbf{Annotation result}                             \\ 
			\hline
			1                & In this paper, we employ \uline{the bounded L-BFGS (Behm et al., 2009) }algorithm for the optimization task, which works well even when the number of weights is large. \par{}\textbf{~}                                 & Model     & L-BFGS (Behm et al., 2009) algorithm\textbf{}            \\
			2                & The evaluation method is \uline{the case insensitive IBM BLEU-4} (Papineni et al., 2002). \par{}\textbf{~}                                                                                                               & Metrics       & BLEU-4\textbf{}                                          \\
			3                & Our \uline{IWSLT data} is the \uline{IWSLT 2009 dialog task data set}. & Dataset       & IWSLT data; IWSLT 2009 dialog task data set\textbf{ }  \\
			4                & We used four datasets: IMDB, Elec, RCV1, and \uline{20-newsgrous (20NG)} to facilitate direct comparison with DL15. \par{}\textbf{~} & {Dataset} & 20-newsgrous (20NG)\textbf{}                                      \\
			5       & Escudero et al.(2000 )used the DSO corpus to highlight the importance of the issue of domain dependence of WSD systems, but did not propose methods such as \uline{active learning} or \uline{countmerging} to address the specific problem of how to \uline{perform domain adaptation for WSD}.                    & Metrics     & active learning, countmerging     \\
			6                & We adapt a \uline{Kneser-Ney language model} to incorporate such growing discounts, resulting in perplexity improvements over \uline{modified Kneser-Ney} and \uline{Jelinek-Mercer} baselines. \par{}\textbf{~} & Model     & Kneser-Ney language model; modified Kneser-Ney; JelinekMercer\textbf{}  \\
			7                & Techniques commonly applied to \uline{automatic text indexation} can be applied to the automatic transcriptions of the broadcast news radio and TV documents. \par{}\textbf{~} & {Task}        & text indexation\textbf{}    \\
			
			\hline
		\end{tabular}
		\caption{\label{table7}Examples for the entity annotation}
	\end{table*}
	
	\subsection{Entity Relation Annotation}
	\textbf{A.3.1 Entity Relation Types}
	
	And for the relations, we have defined the following 11 entity relationships, Table ~\ref{table8} shows examples of correct relation annotations:
	\begin{itemize}
		\item \textbf{measuredBy}: This relationship indicates that the performance of algorithm/model $M_{}$ is typically evaluated using metric $m_{}$.
		\item \textbf{usedFor}:This relationship indicates that the algorithm/model $M_{}$ is typically applied to the NLP task $T_{}$
		\item \textbf{trainedWith}: This relationship indicates that algorithm/model $M_{}$ is pre-trained on dataset $D_{}$.
		\item \textbf{compareWith}: This relationship indicates that authors compare two entities of the same type.
		\item \textbf{enhancedBy}: This relationship indicates that method $M_{1}$ can address its own limitations through method $M_{2}$ to achieve better performance.
		\item \textbf{subclassOf}: This relationship indicates that there is a hypernym-hyponym relationship between the two entities. The hyponym has a narrower scope, while the hypernym has a broader scope. The hyponym inherits the characteristics of the hypernym, and its semantic meaning encompasses or includes that of the hypernym. Additionally, the hyponym possesses unique characteristics of its own.
		\item \textbf{subtaskOf}: This relationship indicates that task $T_{1}$ is a subtask or hyponym of another task $T_{2}$.
		\item \textbf{partOf}: This relationship indicates that entity $E_{1}$ is a component of entity $E_{2}$.
		\item \textbf{evaluatedBy}: This relationship indicates that the NLP task $T_{}$ is typically evaluated using metric $M_{}$
		\item \textbf{evaluatedOn}: This relationship indicates that the NLP task $T_{}$ is typically evaluated on dataset $D_{}$.
		\item \textbf{similarWith}: This relationship indicates that the two entities share the same or highly similar meanings, such as abbreviations and full names of certain met
	\end{itemize}
	\textbf{A.3.2 Annotation Note}
	\begin{itemize}
		\item \textbf{Do not annotate negative relationships}.For example, relationships such as "X is not used in Y" or "X is difficult to apply in Y" should not be annotated.
		\item \textbf{Verify that entities involved in a relationship match the specified types} (e.g., Method-Dataset for TrainedWith). Do not link incorrect entity types through these predefined relationships.
		\item \textbf{Annotate relationships only if there is direct textual evidence or explicit implication}. Avoid inferring relationships that are not clearly stated or strongly suggested in the text.
		\item \textbf{Ensure consistency in relationship annotations across documents}. If uncertain about a relationship, refrain from annotating it.
		\item \textbf{Do not make assumptions about relationships based on personal knowledge or external information}. Rely solely on information explicitly provided in the text.
		\item \textbf{Annotate all identifiable relationships inferred from context, regardless of their quantity}. If contextual clues strongly imply a relationship, annotate it even if mentioned briefly.
	\end{itemize}
	
	\begin{table*}
		\centering
		\fontsize{10}{10}\selectfont
		\renewcommand{\arraystretch}{1.5}
		\begin{tabular}{
				>{\raggedright\arraybackslash}p{0.15\textwidth}
				>{\raggedright\arraybackslash}p{0.3\textwidth}
				>{\raggedright\arraybackslash}p{0.5\textwidth}
			}
			\hline
			\multicolumn{1}{c}{\textbf{Relation type}}& \multicolumn{1}{c}{\textbf{Examples}} &\multicolumn{1}{c}{\textbf{Example sentences}}                                     \\ 
			\hline
			subtaskOf; subtaskOf     
			& \textcolor{purple}{Document-level emotion classification}\(\to\)\textcolor{purple}{classification} ; 
			\textcolor{purple}{sentence-level emotion classification}\(\to\)\textcolor{purple}{classification}    
			&\textcolor{purple}{Document-level emotion classification}, \textcolor{purple}{sentence-level emotion classification} is naturally a multi-label \textcolor{purple}{classification} problem.  \\

			usedFor; usedFor     
			& \textcolor{orange}{SEXTANT}\(\to\)\textcolor{purple}{context extraction};
			\textcolor{orange}{SEXTANT}\(\to\)\textcolor{purple}{grammatical relation extraction};
			& Curran and Moens compared several \textcolor{purple}{context extraction} methods and found that the shallow pipeline and \textcolor{purple}{grammatical relation extraction} used in \textcolor{orange}{SEXTANT} was both extremely fast and produced high-quality results.  \\
			
			usedFor; usedFor; compareWith    
			& \textcolor{orange}{Seq2Seq model}\(\to\)\textcolor{purple}{SQG task} ; 
			\textcolor{orange}{Seq2Seq model}\(\to\)\textcolor{purple}{SQG task} ; 
			\textcolor{purple}{SQG task}---\textcolor{purple}{TQG task}
			& Note that if we directly compare the performance between \textcolor{purple}{SQG task} and \textcolor{purple}{TQG task} under the same model (e.g., the \textcolor{orange}{Seq2Seq model}).  \\
			\hline
		\end{tabular}
		\caption{\label{table8}Examples of the relations annotation}
		
	\end{table*}

\end{document}